\documentclass[review,times,10pt]{elsarticle}

\usepackage{amssymb}
\usepackage{amsmath}
\usepackage{multirow} 
\usepackage{booktabs}
\usepackage{graphicx}
\usepackage{pifont}
\newcommand{\cmark}{\ding{51}} 
\newcommand{\xmark}{\ding{55}}
\usepackage{subcaption}
\newcommand{\subfigwidthreal}{0.11\textwidth}
\usepackage{color}
\usepackage{rotating}

\journal{Patten Recognition}

\begin{document}

\begin{frontmatter}



\title{Learning Multi-scale Spatial-frequency Features \\ for Image Denoising}


\author[NJUST]{Xu~Zhao}
\author[NJU]{Chen~Zhao}
\author[NJUST]{Xiantao~Hu}
\author[NJUST]{Hongliang~Zhang}
\author[NJU]{Ying~Tai \corref{cor}}

\author[NJUST]{Jian~Yang \corref{cor}}

\affiliation[NJUST]{organization={PCA Lab, Key Lab of Intelligent Perception and Systems for High-Dimensional Information of Ministry of Education, School of Computer Science and Engineering},
            addressline={Nanjing University of Science and Technology}, 
            city={Nanjing},
            postcode={210094}, 
            state={Jiangsu},
            country={China}}
            
\affiliation[NJU]{organization={School of Intelligence Science and Technology},
            addressline={Nanjing University}, 
            city={Suzhou},
            postcode={215163}, 
            state={Jiangsu},
            country={China}}
\cortext[cor]{Corresponding author}

\begin{abstract}

Recent advancements in multi-scale architectures have demonstrated exceptional performance in image denoising tasks. 
However, existing architectures mainly depends on a fixed single-input single-output Unet architecture, ignoring the multi-scale representations of pixel level. 
In addition, previous methods treat the frequency domain uniformly, ignoring the different characteristics of high-frequency and low-frequency noise. 
In this paper, we propose a novel multi-scale adaptive dual-domain network (MADNet) for image denoising. 
We use image pyramid inputs to restore noise-free results from low-resolution images.
In order to realize the interaction of high-frequency and low-frequency information, we design an adaptive spatial-frequency learning unit (ASFU), where a learnable mask is used to separate the information into high-frequency and low-frequency components.
In the skip connections, we design a global feature fusion block to enhance the features at different scales.
Extensive experiments on both synthetic and real noisy image datasets verify the effectiveness of MADNet compared with current state-of-the-art denoising approaches.

\end{abstract}



\end{frontmatter}



\section{Introduction}
\label{Introduction}

Image denoising is a crucial task in low-level computer vision \cite{zheng_CVPR, bao2022emotion}.
Due to environmental and equipment limitations, images are often contaminated by various noise during the image acquisition process, which degrades the imaging quality and influences further analysis and processing. 
Early works \cite{BM3D} typically used additive white Gaussian noise (AWGN), which is independent and identically distributed, to create synthetic noise images.
However, noise in real images is complex, while traditional optimization algorithms are generally time-consuming or unable to directly eliminate real noise. 

\begin{figure}[t!]
\centering
\includegraphics[width=\textwidth]{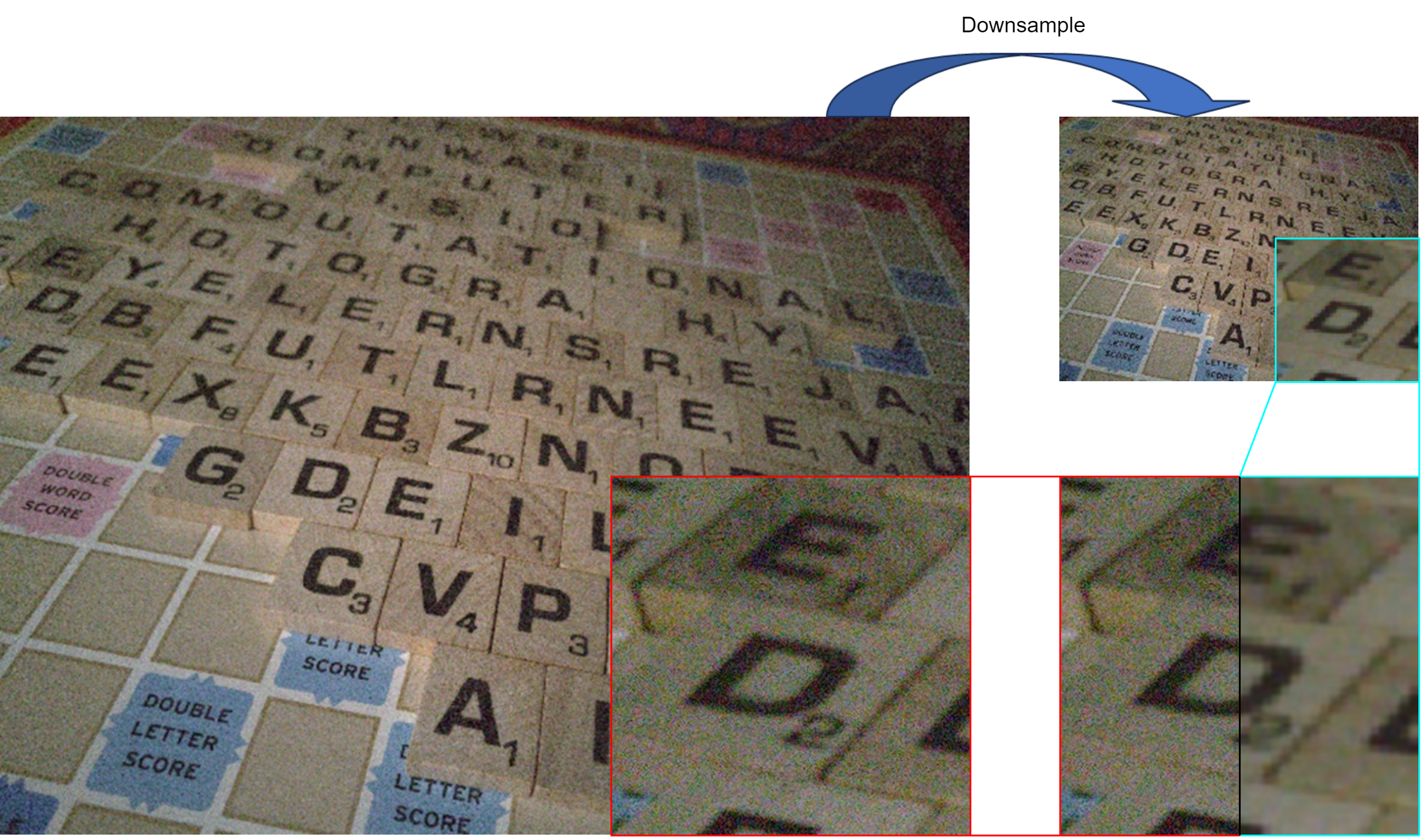}
\caption{Comparison of degradation magnitude of low-resolution images. The observed degradation in low-resolution images provides cleaner contextual information for network learning, highlighting the importance of incorporating multi-scale representations in architecture design.}
\label{downsample.png}
\end{figure}

Learning-based methods \cite{xie2024node, ning2024enhancement} remove noise by training models to establish mappings from noisy images to clean ones. As deep neural networks gain popularity, various architectures have been developed.
CNNs have achieved remarkable success \cite{ji2020real} thanks to their large modeling capacities and substantial progress in network training and design methodologies.
For instance,
DnCNN \cite{DnCNN} introduces residual learning and batch normalization to remove noise.
Despite numerous advances in image denoising, challenges such as the small receptive field of conventional CNNs still exist, thus preventing it from modeling long-range pixel dependencies.
Transformer models \cite{Restormer, SUNet, ning20253d} have recently been applied to image denoising tasks for their effectiveness in capturing long-range pixel interactions.
Restormer \cite{Restormer} introduces an efficient design featuring channel-based self-attention and a gated deepwise-convolution feed-forward network.
SUNet \cite{SUNet} considers Swin Transformer \cite{liu2021swin} as main backbone and integrate it into the UNet architecture.

Several recent methods \cite{CFPNet, yan_TIM} explored multi-scale representations, achieving good results.
However, these works design their network architectures mainly rely on a \textit{fixed single-input single-output Unet architecture}, ignoring the multi-scale representations of pixel level. 
In Figure~\ref{downsample.png}, we show how to extract cleaner version from the low-resolution noisy images using multi-scale representations. We observe that noise levels reduce significantly in low-resolution images.
This suggests that the low-resolution images suffered from less degradation (see Table~\ref{tab:performance_metrics}), providing the network with \textit{noiseless and contextual information}. 
Hence, these architectures should also make full consideration of low-resolution images during the architecture design.

\begin{table}[t!]
\centering
\tabcolsep=0.6cm
\begin{tabular}{cccc}
\toprule
\textbf{Scale} & \textbf{MSE} & \textbf{PSNR} & \textbf{SSIM} \\
\midrule
1.0    & $2.9 \times 10^{-4}$ & 35.23 & 0.83 \\
0.5    & $1.7 \times 10^{-4}$ & 37.61 & 0.90 \\
0.25   & $1.6 \times 10^{-4}$ & 37.74 & 0.92 \\
0.125  & $1.5 \times 10^{-4}$ & 38.06 & 0.94 \\
\bottomrule
\end{tabular}
\caption{Performance metrics at different scales.}
\label{tab:performance_metrics}
\end{table}

\begin{figure}[t!]
\centering
\includegraphics[width=\textwidth]{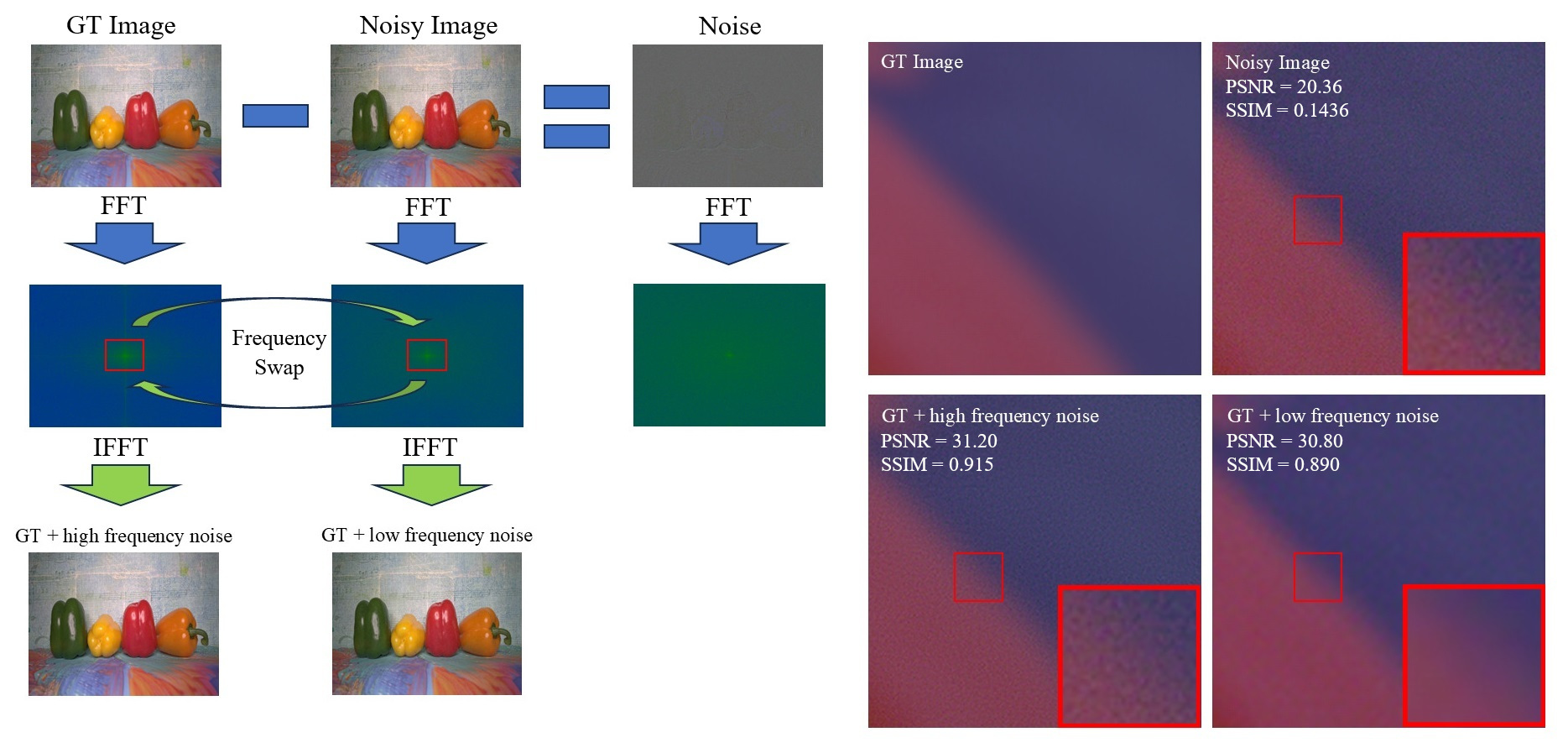}
\caption{Comparison of degradation magnitude of low-resolution images. The degradation of the image is different for high-frequency noise and low frequency noise.}
\label{frequency.jpg}
\end{figure}

While multi-scale frameworks excel at capturing hierarchical features, they often rely on spatial-domain operations that struggle to isolate task-relevant frequency components (e.g., high-frequency textures vs. low-frequency semantic structures). To address this, MADNet introduces an adaptive frequency mask that explicitly separates and refines frequency bands across scales, enabling task-driven optimization.
Some studies \cite{FADNet, Fourmer} have reported employing frequency domain transformations, such as the Fast Fourier Transform (FFT), in deep learning for image denoising.
FADNet \cite{FADNet} combines Unet with dynamic filters and the Fourier transform.
Fourmer \cite{Fourmer} uses fourier spatial modeling and fourier channel evolution to disentangle image degradation and content.
However, these FFT-based methods treat the frequency domain uniformly and do not consider the difference between high-frequency and low-frequency information. 
As shown in Figure~\ref{frequency.jpg}, we found that different types of degradation can affect the image content in different degrees. 
The hybrid images were generated by swapping low-frequency components between the clean and degraded images in the Fourier domain. This produced two versions: one with high-frequency noise (retaining low-frequency components from the clean image) and another with low-frequency noise (retaining high-frequency components from the clean image).
It can be seen that the degradation of the image is different for high-frequency noise and low-frequency noise. 
Therefore, it is necessary to \textit{deal with high-frequency information and low-frequency information separately}.

In this paper, we introduce MADNet, a multi-scale adaptive dual-domain network designed to tackle key challenges in image denoising.
Specifically, MADNet extends the Unet architecture with dual-domain modulation layers (DDML) and a global feature fusion block (GFFB).
For DDML, a novel adaptive spatial-frequency learning unit (ASFU) is introduced to capture long-range dependencies in both spatial and frequency domains.
The ASFU employs a learnable adaptive mask to separate high-frequency and low-frequency components, followed by transposed self-attention \cite{Restormer} for long-range channel dependency modeling.
The adaptive rectangular binary mask is parameterized by a learnable coefficient, which scales its dimensions proportionally to the input size. This allows automatic optimization of frequency separation during training.
GFFB concatenated features at different scales to obtain global feature, which is used to enhance features from different scales through transposed self-attention.

In conclusion, the main contributions of our method are as follows:

\noindent (1) We propose a novel multi-scale adaptive dual-domain network for efficient noise removal, which enhance the features at different scales using multi-input multi-output architecture with multi-scale image contents and a global feature fusion block.

\noindent (2) We propose the adaptive spatial-frequency learning unit (ASFU), which uses a learnable mask to separate high-frequency and low-frequency components and  captures long-range channel dependencies via transposed self-attention. 

\noindent (3) We conduct experiments on several synthetic image datasets and real image datasets. The results demonstrate that our model achieves state-of-the-art performances on both synthetic and real noisy images.

\section{Related Works}
\label{Related Works}

\subsection{Image Denoising}
\label{Image Denoising}
Low-level has received a lot of attention due to its wide range of applications~\cite{Tracking, ning2024occluded,chen2024transformer,hu2025exploiting,wang2025not,wang2025dornet,wu2024deep}, and image denoising is one of the important tasks.

Image denoising methods can generally be divided into two main categories: prior-based methods and learning-based methods.
Prior-based methods are based on some priors from natural images.
For instance, 
BM3D \cite{BM3D} uses nonlocal similarity to group similar blocks into a 3D matrix and applies collaborative filtering to eliminates noisy pixels.
These methods are usually time-consuming and hard to remove complex real noise.
%
Different from prior-based methods that heavily rely on handcrafted priors, the learning-based methods \cite{xie2024node, 
zhang_PR} focus on learning a latent mapping from the noisy image to the clean version. 
Recently, deep learning-based methods have achieved remarkable progress.
For instance,
DeamNet \cite{DeamNet} introduces traditional methods into deep neural networks to enhance the interpretability of the model.
NIFBGDNet \cite{NIFBGDNet} uses the features of a negative image to guide the denoising process.
DRANet \cite{DRANet} uses dual network structure to combine different image features.
Despite CNNs provide both efficiency and generalization ability, the limited receptive field of the convolutional operator restricts its ability to model long-range pixel dependencies.
Recent methods expand the receptive field through multi-scale architectures, but they do not effectively overcome this limitation.

Transformer, originally developed for sequence processing in natural language tasks, have gained traction in the low-level vision field \cite{an_AAAI2024, Restormer, ning2023pedestrian} due to their ability to capture long-range dependencies through global modeling.
For instance,
Uformer \cite{Uformer} integrates a Unet architecture with locally-enhanced window Transformer blocks and a multi-scale restoration modulator. 
Restormer \cite{Restormer} introduces an efficient design featuring channel-based self-attention and a gated deepwise-convolution feed-forward network. 
However, recent works are limited to single-input single-output architecture, ignoring the multi-scale representations of images.

Different from the aforementioned methods, our work uses image pyramid inputs to provide \textit{noiseless-robustness and contextual information}, exploiting feature correlations in different image scale spaces.

\subsection{Frequency learning}
\label{Frequency learning}

Recent years, many methods \cite{zhao2024cycle, hu_2023_tcsvt} have extracted information from the frequency domain to address various low-level vision problems.
For instance,
\cite{fourier_loss} suggests a supervision loss within the Fourier domain in low-level generative tasks.
SFGNet \cite{SFGNet} integrates spatial-frequency interaction and gradient maps through a two-stage process of dense spatial-frequency fusion and gradient-aware correction. 
FADNet \cite{FADNet} employs two subnetworks: a UNet structure for noise estimation and a frequency domain network to train the adaptive convolutional layer.
WF-Diff \cite{WF-Diff} uses frequency domain information and diffusion models to enhance and refine frequency details.
CFPNet \cite{CFPNet} uses cosine transforms to segment noise frequencies and fuses the signals to enhance high-frequency details.

Despite the wide exploration of various problems, existing works treat the frequency domain uniformly. In contrast, we propose an adaptive frequency enhancement block (AFEB) that \textit{adaptively separates high-frequency and low-frequency information}, effectively capturing their distinct characteristics.

\section{Method}
\label{Method}

In this section, we propose a multi-scale adaptive dual-domain network (MADNet) for image denoising. The overall architecture of the proposed method is shown in Figure~\ref{structure.jpg}. In the following, we introduce the architecture principles of MSANet and provide a detailed explanation of each module.

\subsection{Overall Network Structure}
\label{Overall Network Structure}

\begin{figure}[t!]
\centering
\includegraphics[width=\textwidth]{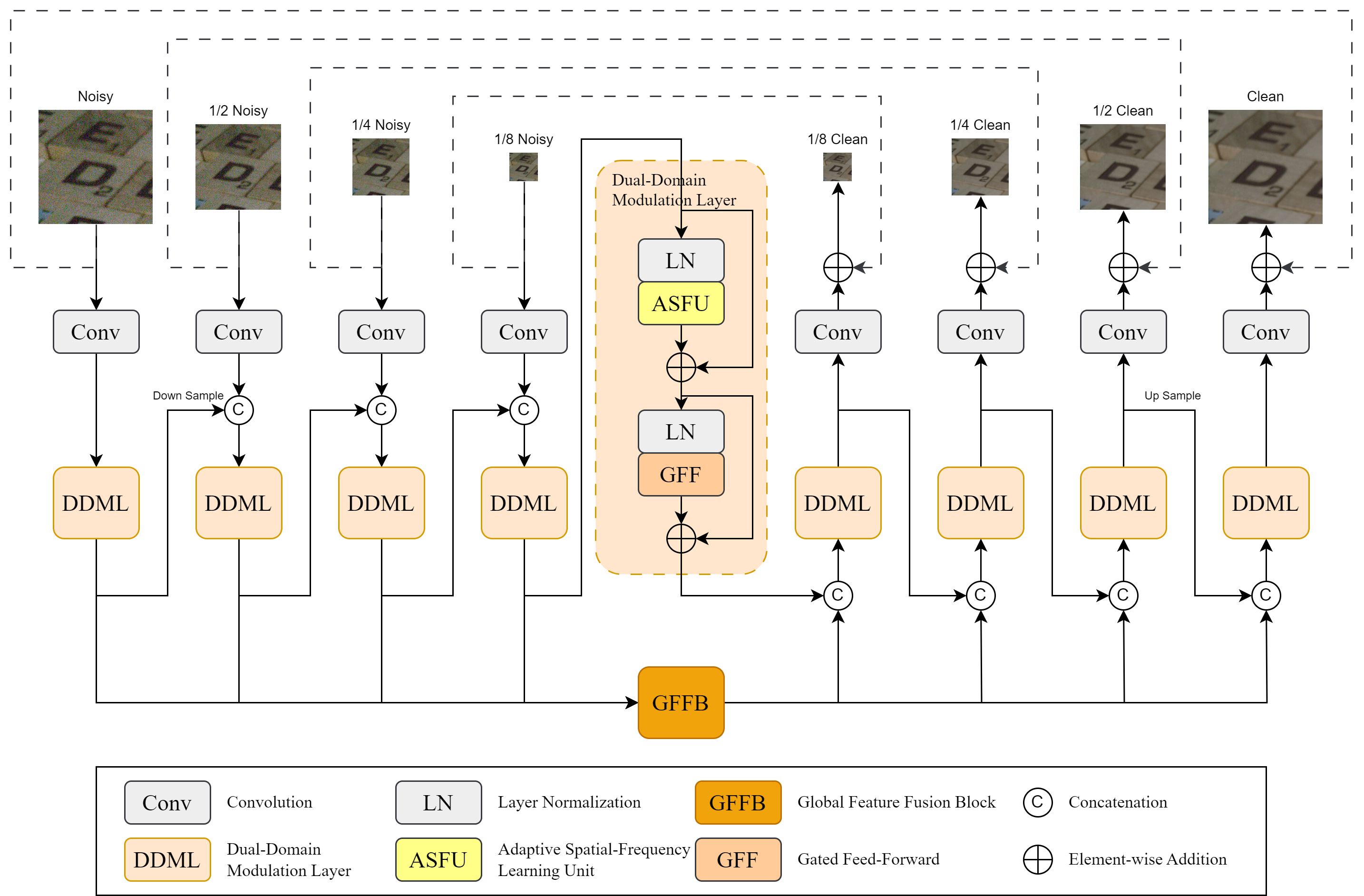}
\caption{The structure of the MADNet. A Unet structure processes multi-scale pyramid inputs with dual-domain modulation layers (DDMLs) and global feature fusion blocks (GFFBs) to extract and refine features, producing residual images and clear outputs at multiple scales.}
\label{structure.jpg}
\end{figure}

As illustrated in Figure~\ref{structure.jpg}, given a noisy image $x \in \mathbb{R}^{H \times W \times 3}$, MADNet first produces images pyramid inputs using the interpolation operator to down-sample the original image into multiple scales. The network takes these images as multi-inputs and extracts the shallow features $\{F_i, i = 0,1,2,3\} \in \{\mathbb{R}^{\frac{H}{n} \times \frac{W}{n} \times nC}, n = 1, 2, 4, 8\}$ using a shallow $3 \times 3$ convolutional layer.
Next, the shallow features $F_0$ are successively passed through a 4-stage hierarchical encoder-decoder structure. 
Based on the initial features from each scale, dual-domain modulation layers (DDMLs) perform deep feature extraction by adaptive spatial-frequency learning units (ASFUs) and gated feed-forward networks (GFFs). The low-resolution shallow feature is then concatenated with the deep feature.
The deep features at each scale are concatenated via a global feature fusion block (GFFB) in skip connections to maintain structural and textural details. 
Finally, a $3 \times 3$ convolution layer is applied to the deep features, generating the residual image $\{s_i, i = 0,1,2,3\} \in \{\mathbb{R}^{\frac{H}{n} \times \frac{W}{n} \times C}, n = 1, 2, 4, 8\}$ to obtain the clear image $\{\hat{x}_i, i = 0,1,2,3\} \in \{\mathbb{R}^{\frac{H}{n} \times \frac{W}{n} \times C}, n = 1, 2, 4, 8\}$ at each scale.

\subsection{Dual-Domain Modulation Layer}
\label{Dual-Domain Modulation Layer}

The proposed dual-domain modulation layer (DDML) effectively integrates shallow and deep features. As illustrated in Figure 3, the input feature $ F_i $ is normalized via layer normalization (LN) and refined through the ASFUs to extract initial feature representations.
Subsequently, the aggregated features undergo layer normalization and are refined by the gated feed-forward network (GFF) to integrate multi-scale and dual-domain information, formulated as follows:

\begin{equation}
\label{eq:DDML}
F'_i = \text{GFF}(\text{LN}(\text{SFB}(\text{LN}(F_i)) + F_i)) + F_i, 
\end{equation}
where $F_i$ is the input feature, $F'_i$ is the output feature, $\text{LN}$ denotes the layer normalization.
%
The structure of GFF is shown in Figure~\ref{modules.jpg}. Specifically, the input feature is processed by a standard convolution followed by a depthwise convolution (DWconv). The output of the depthwise convolution is split into two branches, then applies a gating mechanism through an element-wise multiplication. 

\subsection{Adaptive Spatial-Frequency Learning Unit}
\label{adaptive spatial-frequency learning unit}

\begin{figure}[t!]
\centering
\includegraphics[width=\textwidth]{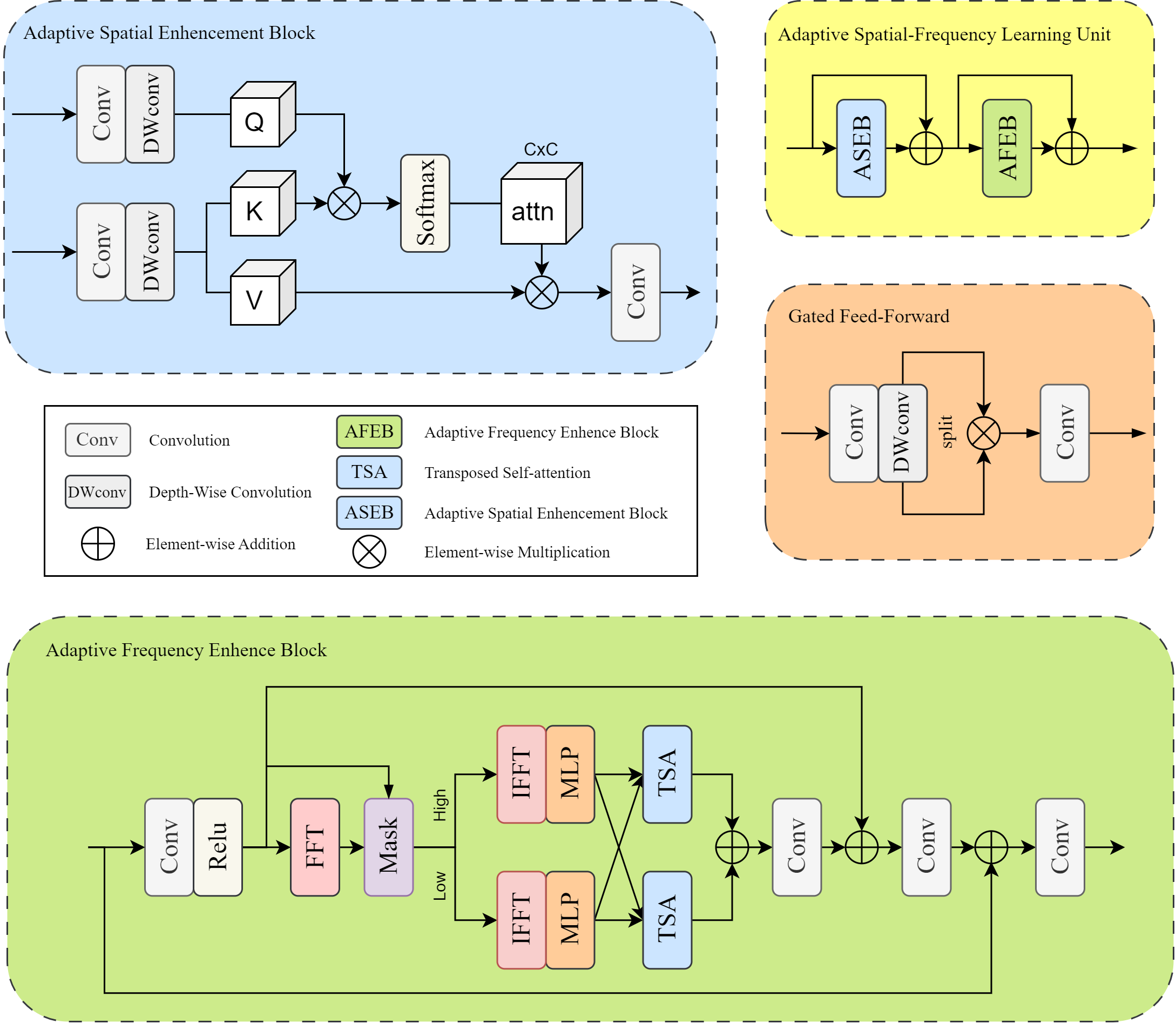}
\caption{Detailed architecture of each block in MADNet.}\label{modules.jpg}
\end{figure}

We show the structure of ASFU in Figure \ref{modules.jpg}, which has an adaptive frequency enhancement block (AFEB) and a transposed self-attention as our adaptive spatial enhencement block (ASEB) for interaction of dual domain representations.
Here, we review the Fourier transform. Given an image $x \in \mathbb{R}^{H \times W \times C}$, the fast Fourier transform (FFT) $\mathcal{F}$ converts it to frequency space as the complex component $\mathcal{F}(x)$: 
\begin{equation}
\label{eq:FFT}
\mathcal{F}(x)(u, v) = \frac{1}{\sqrt{HW}}\sum_{h=0}^{H-1}\sum_{w=0}^{W-1}x(h, w)\,e^{-\,j\,2\pi\!\left(\frac{h}{H}u \;+\; \frac{w}{W}v\right)},
\end{equation}
where $u$ and $v$ are the coordinates of the Fourier space. $\mathcal{F}^{-1}(x)$ represents the inverse fast Fourier transform (IFFT).

\subsubsection{Adaptive Frequency Enhancement Block}
\label{adaptive frequency enhancement block}

AFEB aims to effectively extract and enhance high-frequency and low-frequency information adaptively. The input feature $F_i$ is first processed through a convolution layer followed by a ReLU activation function to extract basic spatial features. These features are then transformed into the frequency domain using the FFT. Then a learnable adaptive frequency mask is applied to separate the high-frequency and low-frequency components:
\begin{equation}
F_{\text{freq}} = \text{FFT}(\text{Relu}(\text{Conv}(F_i))),
\end{equation}
\begin{equation}
F_{\text{high}}, F_{\text{low}} = Mask(F_{\text{freq}}).
\end{equation}

For each frequency component, we apply the MLP after the IFFT to refine the features. 
These components are further enhanced through transposed self-attention (TSA) to emphasize critical feature channels adaptively.
The outputs of both branches are then combined via element-wise addition:
\begin{equation}
F_{\text{high}}' = \text{TSA}(\text{MLP}(\text{IFFT}(F_{\text{high}}))), \quad
F_{\text{low}}' = \text{TSA}(\text{MLP}(\text{IFFT}(F_{\text{low}}))),
\end{equation}
\begin{equation}
F_{\text{combined}} = F'_{\text{high}} + F'_{\text{low}},
\end{equation}
where $ F_{\text{high}} $ and $ F_{\text{low}} $ denote the high- and low-frequency components separated from the frequency-domain representation $ F_{\text{freq}} $, $ F_{\text{high}}' $ and $ F_{\text{low}}' $ are their refined spatial-domain counterparts after IFFT, MLP, and TSA operations, and $ F_{\text{combined}} $ is further processed by a sequence of convolution operations and residual connections to integrate the enhanced frequency-domain information.

\subsubsection{Adaptive Spatial Enhencement Block}
\label{transposed self-attention}

The transposed self-attention (TSA) serves as the adaptive spatial enhancement block (ASEB) for adaptive refinement of spatial features.
TSA module is designed to capture long-range dependencies across channels. As is shown in Figure~\ref{modules.jpg}, the feature extraction process is defined as:
\begin{equation}
    F^* = \text{Softmax} \left( \frac{QK^\top}{\alpha} \right) V,
\end{equation}
\begin{equation}
    Q = W^Q_d W^Q_p (F^*), \quad K = W^K_d W^K_p (F^*), \quad V = W^V_d W^V_p(F^*),
\end{equation} 
where $* \in \{l, h\}$ is an indicator for low/high frequency, $W^{(\cdot)}_p$ is the $1 \times 1$ point-wise convolution and $W^{(\cdot)}_d$ is the $3 \times 3$ depth-wise convolution, and $Q$, $K$, and $V$ are query, key, and value projections. Additionally, $\alpha$ is a learnable scaling factor to control the magnitude of the dot product result of $Q$ and $K$ before using the softmax function.

\begin{figure}[htbp]
    \centering
    \begin{subfigure}{0.19\textwidth}
        \includegraphics[width=\linewidth]{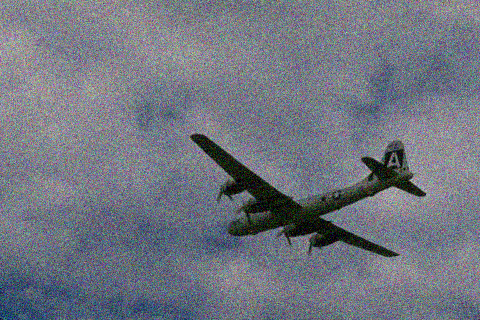}
        \caption*{(a) Noisy Image}
    \end{subfigure}
    \hfill
    \begin{subfigure}{0.19\textwidth}
        \includegraphics[width=\linewidth]{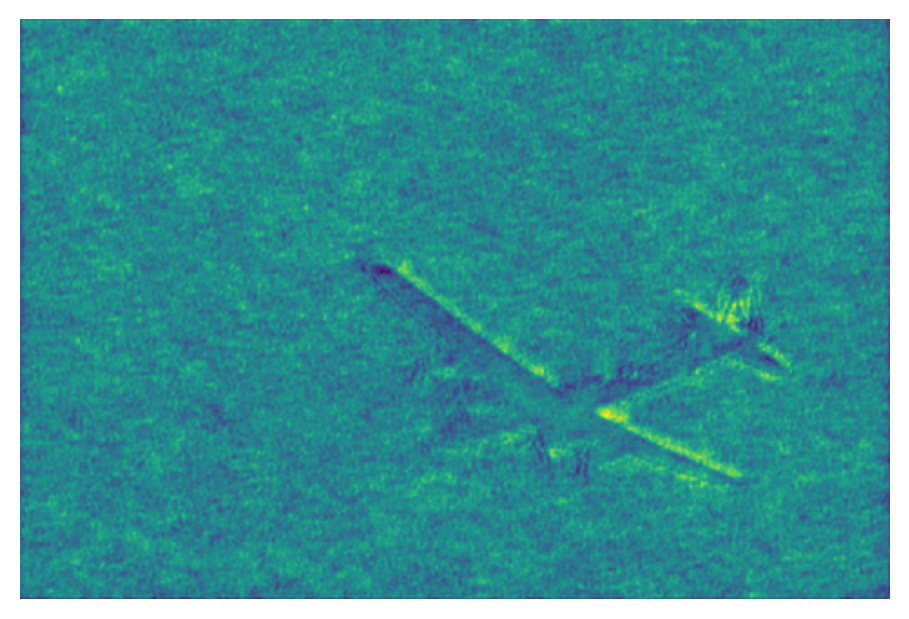}
        \caption*{(b) Fusion Layer 1}
    \end{subfigure}
    \hfill
    \begin{subfigure}{0.19\textwidth}
        \includegraphics[width=\linewidth]{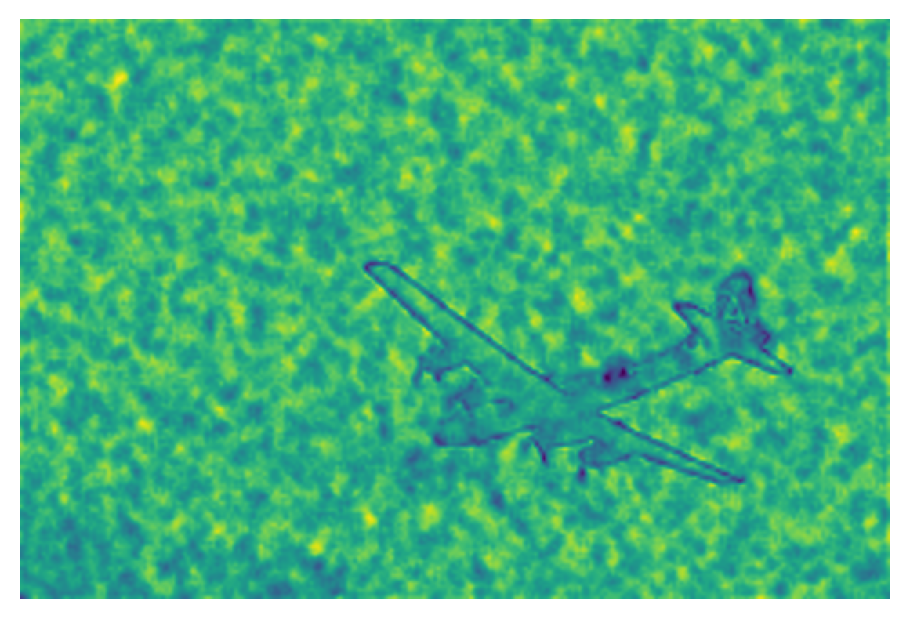}
        \caption*{(c) Fusion Layer 2}
    \end{subfigure}
    \hfill
    \begin{subfigure}{0.19\textwidth}
        \includegraphics[width=\linewidth]{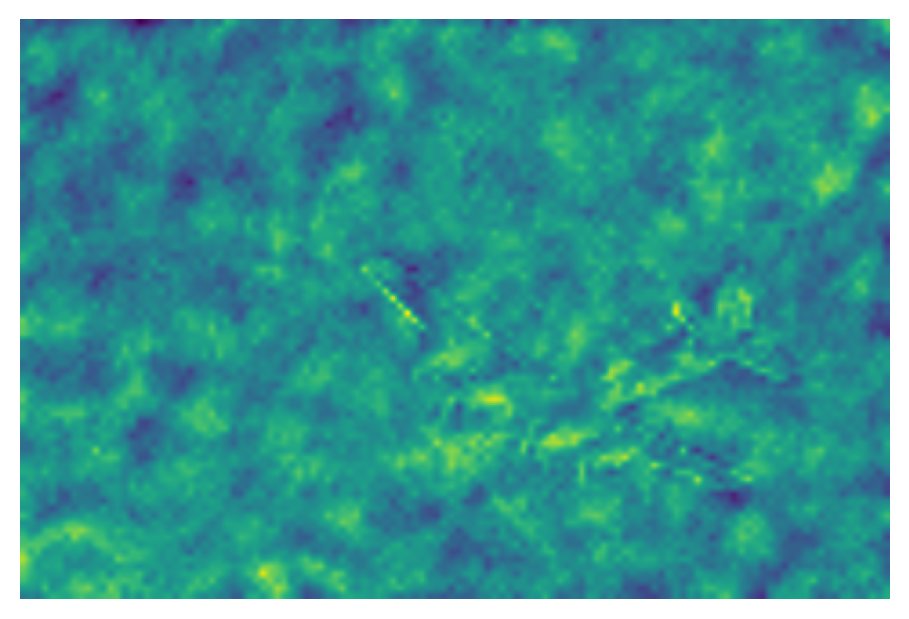}
        \caption*{(d) Fusion Layer 3}
    \end{subfigure}
    \hfill
    \begin{subfigure}{0.19\textwidth}
        \includegraphics[width=\linewidth]{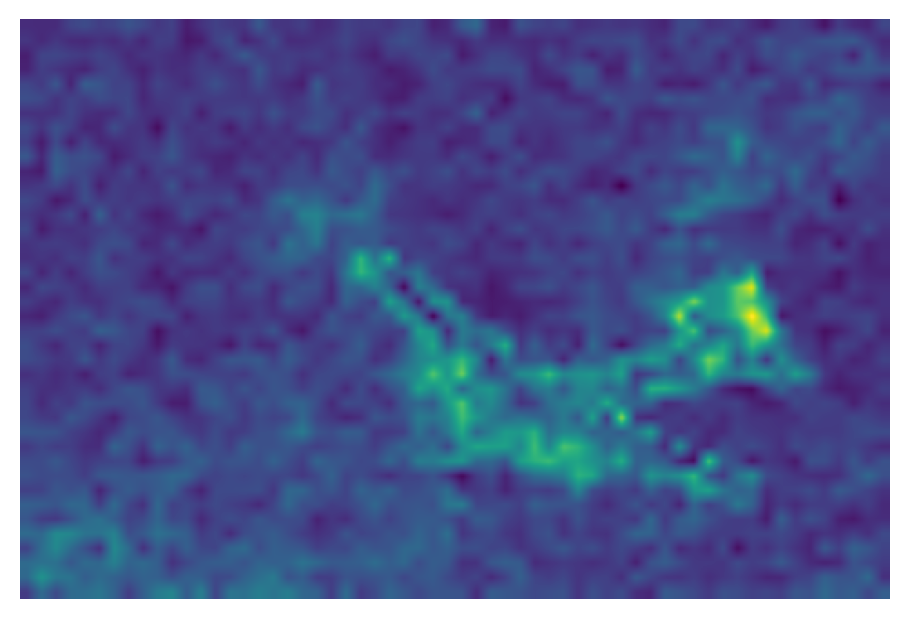}
        \caption*{(e) Fusion Layer 4}
    \end{subfigure}

    \caption{Feature visualization across fusion layers: (a) Noisy image (b)-(e) Fusion Layer feature heatmaps.}
    \label{fig:feature_visualization}
\end{figure}

\subsection{Global Feature Fusion Block}
\label{Global Feature Fusion Block}

The GFFB unifies diverse feature representations into a single global feature. Features from multiple inputs are concatenated along the channel dimension to produce a unified global representation, and then adaptively enhance each input feature at different scales using transposed self-attention, which is expressed as:
\begin{align}
F_g &= \text{Concat}(F_1, F_2, F_3, F_4), \\
\hat{F_i} &= \text{ASEB}(F_g, F_i), \quad i = 1,2,3,4,
\end{align}
where $F_1, F_2, F_3, F_4$ are the input features from different scales. $F_g$ is the unified global feature obtained by concatenating the input features along the channel dimension. $\text{ASEB}(\cdot)$ denotes the adaptive spatial enhencement block, and $\hat{F_i}$ represents the refined feature.

We show feature heatmaps in Figure \ref{fig:feature_visualization}, which illustrates the attention distribution across four hierarchical scales (1×, 0.5×, 0.25×, and 0.125× resolutions). These visualizations demonstrate that finer scales emphasize local details, while coarser scales prioritize global semantic structures.

\subsection{Loss Function}
\label{Loss function}

In order to supervise the learning process of the network, we choose a spatial domain loss and a frequency domain loss as training objectives to guide the optimization. In the spatial domain, we apply the multi-scale Charbonnier loss \cite{Charbonnier_loss} as: 
\begin{equation}
\label{eq:loss_s}
\mathcal{L}_s = \sum_{s=1}^{S} \alpha_s \left( \frac{1}{N_s} \sum_{i=1}^{N_s} \sqrt{\bigl(r_{i}^{(s)} - t_{i}^{(s)}\bigr)^2 + \epsilon^2} \right),
\end{equation}
where \(S\) denotes the total number of scales, \(N_s\) is the number of samples (or pixels) at scale \(s\). The terms \(r_{i}^{(s)}\) and \(t_{i}^{(s)}\) represent the predicted and ground truth values at the \(i\)-th sample of scale \(s\), respectively. The term \(\epsilon\) is a small constant  for numerical stability, \(\alpha_s\) is a weighting factor reflecting the importance of each scale.
In the frequency domain, the loss is formulated as:
\begin{equation}
\label{eq:loss_f}
\mathcal{L}_f = \sum_{s=1}^{S} \alpha_s \left( \frac{1}{N_s} \sum_{i=1}^{N_s} \Bigl\lVert \mathcal{F}\bigl(r^{(s)}\bigr) - \mathcal{F}\bigl(t^{(s)}\bigr)\Bigr\rVert_{1} \right),
\end{equation}
where \(\mathcal{F}\) refers to the Fast Fourier transform, with \(\lVert \cdot \rVert_1\) denoting the \(L_1\) norm. 
The overall loss function is designed as:
\begin{equation}
\label{eq:loss}
\mathcal{L} = \mathcal{L}_s + \mathcal{L}_f. 
\end{equation}

\section{Experiments}
\label{Experiments}

In this section, we first introduce the synthetic and real noise datasets and the experimental settings, and then show the quantitative and qualitative results on these datasets. Finally, we perform analysis experiments including ablation study and model complexity analysis. In tables, the best quality scores of the evaluated methods are \textbf{highlighted}.

\subsection{Dataset}
\label{Dataset}
Training and testing are conducted on both real and synthetic noise datasets. The Flickr2K dataset \cite{Flick2K} is used for synthetic data, with training and testing performed on both color and grayscale images. The dataset comprises 2,650 high-resolution color images. Synthetic noisy patches are generated by adding additive white Gaussian noise (AWGN) with noise levels ranging from (0, 50] to clean patches. For evaluation, the BSD68 \cite{BSD68_Set12} and Set12 \cite{BSD68_Set12} datasets are selected, with performance assessed at noise levels 15, 25, 30, and 50. The SIDD Medium dataset \cite{SIDD} is employed for training on real data, providing 320 pairs of high-resolution noisy images and their nearly noise-free counterparts. For evaluation, the SIDD test set \cite{SIDD} and the DND sRGB \cite{DND} test set are selected. High-resolution images are randomly cropped into 128×128 patches, with data augmentation performed via rotation and flipping operations.

\subsection{Experimental setting}
\label{Experimental setting}

The parameters of our models were optimized using the Adam optimizer.
For synthetic noise, we train for $6 \times 10^5$ iterations with an initial learning rate of 
$1 \times 10^{-4}$, halving the learning rate after every $1 \times 10^5$ iterations. 
For real noise, we train for 150 epochs with an initial learning rate of 
$2 \times 10^{-4}$, gradually decreasing it to $1 \times 10^{-6}$ following a 
cosine annealing schedule.

\subsection{Results on AWGN color image denoising}
\label{Results on AWGN color image Denoising}

We evaluated the denoising performance of the MADNet on the synthetic noisy images. The DudeNet \cite{DudeNet}, DeamNet\cite{DeamNet}, SwinIR\cite{SwinIR}, MSANet\cite{MSANet}, AirNet \cite{AirNet}, MWDCNN \cite{MWDCNN}, and DRANet\cite{DRANet} were compared.

\begin{table}[ht]
\centering
\scalebox{0.58}{
\begin{tabular}{c l c c c c c c c c c c c c}
\toprule
\multirow{2}{*}{\textbf{Datasets}} & \multirow{2}{*}{\textbf{Methods}} 
& \multicolumn{3}{c}{$\sigma = 15$} 
& \multicolumn{3}{c}{$\sigma = 25$} 
& \multicolumn{3}{c}{$\sigma = 30$}
& \multicolumn{3}{c}{$\sigma = 50$} \\
\cmidrule(lr){3-5}\cmidrule(lr){6-8}\cmidrule(lr){9-11}\cmidrule(lr){12-14}
& & \textbf{PSNR} & \textbf{SSIM} & \textbf{LPIPS} & \textbf{PSNR} & \textbf{SSIM} & \textbf{LPIPS} & \textbf{PSNR} & \textbf{SSIM} & \textbf{LPIPS} & \textbf{PSNR} & \textbf{SSIM} & \textbf{LPIPS} \\
\midrule
\multirow{8}{*}{CBSD68}
 & DudeNet\cite{DudeNet}  & 32.78 & 0.896 & 0.0883 & 29.71 & 0.814 & 0.1101 & 28.94 & 0.7797 & 0.1502 & 25.06 & 0.592 & 0.2900 \\
 & DudeNet\cite{DudeNet}  & 32.78 & 0.896 & 0.0883 & 29.71 & 0.814 & 0.1101 & 28.94 & 0.7797 & 0.1502 & 25.06 & 0.592 & 0.2900 \\
 & DeamNet\cite{DeamNet}  & 34.10 & 0.932 & 0.0530 & 31.48 & 0.889 & 0.0918 & 30.60 & 0.8695 & 0.1102 & 28.28 & 0.804 & 0.1732 \\
 & SwinIR\cite{SwinIR}    & 33.97 & 0.929 & 0.0662 & 31.32 & 0.883 & 0.1202 & 30.42 & 0.8621 & 0.1472 & 28.05 & 0.790 & 0.2356 \\
 & MSANet\cite{MSANet}   & 34.10 & 0.932 & 0.0532 & 31.51 & 0.889 & 0.0970 & 30.64 & 0.8702 & 0.1067 & 28.32 & 0.804 & 0.1840  \\
 & AirNet\cite{AirNet}   & 34.14 & 0.933 & 0.0529 & 31.48 & 0.889 & 0.0918 & 30.58 & 0.8685 & 0.1125 & 28.23 & 0.802 & 0.1830  \\
 & MWDCNN\cite{MWDCNN}   & 34.01 & 0.931 & 0.0545 & 31.38 & 0.886 & 0.0970 & 30.49 & 0.8663 & 0.1180 & 28.14 & 0.797 & 0.1936 \\
 & DRANet\cite{DRANet}   & 34.18 & 0.933 & 0.0523 & 31.56 & 0.890 & 0.0942 & 30.68 & 0.8706 & 0.1137 & 28.37 & 0.806 & 0.1829 \\

 & \textbf{MADNet}        & \textbf{34.21} & \textbf{0.933} & \textbf{0.0522} & \textbf{31.59} & \textbf{0.891} & \textbf{0.0928} & \textbf{30.72} & \textbf{0.8716} & \textbf{0.1120} & \textbf{28.41} & \textbf{0.807} & \textbf{0.1774} \\
\midrule
\multirow{8}{*}{Set12}
 & DudeNet\cite{DudeNet}  & 33.94 & 0.904 & 0.0974 & 30.76 & 0.825 & 0.1281 & 30.04 & 0.7977 & 0.1455 & 25.73 & 0.602 & 0.3197 \\
 & DeamNet\cite{DeamNet}  & 35.46 & 0.937 & 0.0575 & 33.01 & 0.907 & 0.0993 & 32.16 & 0.8944 & 0.1169 & 29.82 & 0.853 & 0.1691 \\
 & SwinIR\cite{SwinIR}    & 35.25 & 0.933 & 0.0865 & 32.78 & 0.901 & 0.1361 & 31.92 & 0.8882 & 0.1560 & 29.49 & 0.842 & 0.2179 \\
 & MSANet\cite{MSANet}   & 35.49 & 0.937 & 0.0518 & 33.07 & 0.907 & 0.0933 & 32.22 & 0.8945 & 0.1108 & 29.90 & 0.853 & 0.1735 \\
 & AirNet\cite{AirNet}   & 35.55 & 0.937 & 0.0515 & 33.10 & 0.908 & 0.1212 & 32.30 & 0.8960 & 0.1200 & 29.83 & 0.854 & 0.2501 \\
 & MWDCNN\cite{MWDCNN}   & 35.32 & 0.935 & 0.0648 & 32.86 & 0.905 & 0.1064 & 32.01 & 0.8919 & 0.1236 & 29.60 & 0.847 & 0.1795 \\
 & DRANet\cite{DRANet}   & 35.61 & 0.939 & 0.0608 & 33.14 & 0.908 & 0.1046 & 32.29 & 0.8957 & 0.1210 & 29.96 & 0.855 & 0.1734 \\

 & \textbf{MADNet}        & \textbf{35.64} & \textbf{0.938} & \textbf{0.0618} & \textbf{33.18} & \textbf{0.908} & \textbf{0.1062} & \textbf{32.32} & \textbf{0.8953} & \textbf{0.1221} & \textbf{29.99} & \textbf{0.855} & \textbf{0.1673} \\
\bottomrule
\end{tabular}
}
\caption{Quantitative results of the AWGN color noise removal evaluation for different noise levels $\sigma$.}
\label{tab:synthetic_comparison_color}
\end{table}

Table \ref{tab:synthetic_comparison_color} provides a comprehensive comparison on the color images at three noise levels ($\sigma$ = 15, 25, 50).
As shown in the CBSD68 quantitative evaluations, our method consistently achieves the highest PSNR values across all noise levels, outperforming competing methods by maintaining superior quantitative metrics.
Specifically, at $\sigma$ = 15, our model achieves a PSNR of 34.21 dB and an SSIM of 0.933. As the noise level increases to $\sigma$ = 50, our model maintains a PSNR of 28.41 dB and an SSIM of 0.807, reflecting its effectiveness in preserving image details under severe noise conditions. 
A consistent performance advantage is observed on the Set12 benchmark, where our method achieves superior PSNR values across all noise levels and maintains 0.855 SSIM at $\sigma$ = 50.

\begin{figure}[htbp]
    \centering
    \begin{subfigure}{\subfigwidthreal}
        \includegraphics[width=\linewidth]{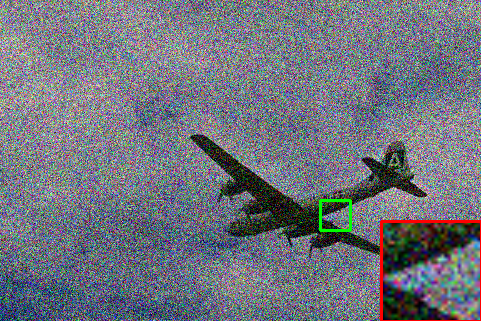}
        \caption*{Noisy}
    \end{subfigure}
    \begin{subfigure}{\subfigwidthreal}
        \includegraphics[width=\linewidth]{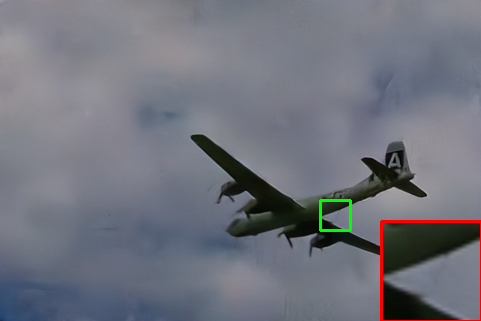}
        \caption*{0.9541}
    \end{subfigure}
    \begin{subfigure}{\subfigwidthreal}
        \includegraphics[width=\linewidth]{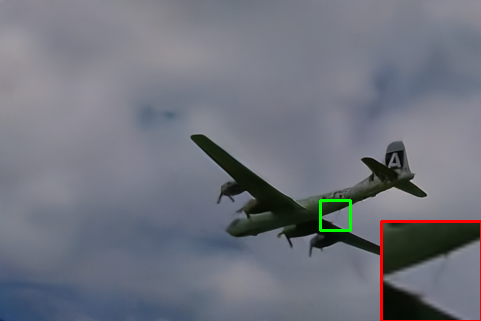}
        \caption*{0.9629}
    \end{subfigure}
    \begin{subfigure}{\subfigwidthreal}
        \includegraphics[width=\linewidth]{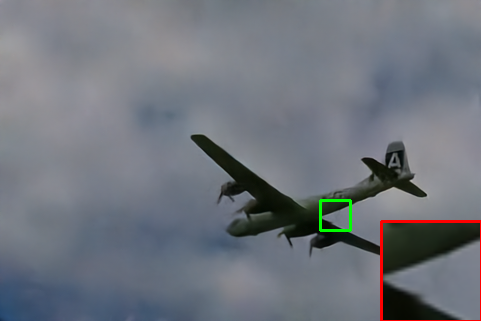}
        \caption*{0.9571}
    \end{subfigure}
    \begin{subfigure}{\subfigwidthreal}
        \includegraphics[width=\linewidth]{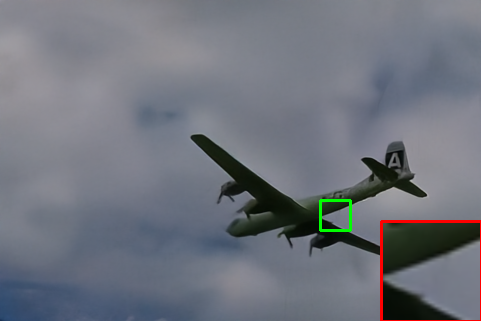}
        \caption*{0.9615}
    \end{subfigure}
    \begin{subfigure}{\subfigwidthreal}
        \includegraphics[width=\linewidth]{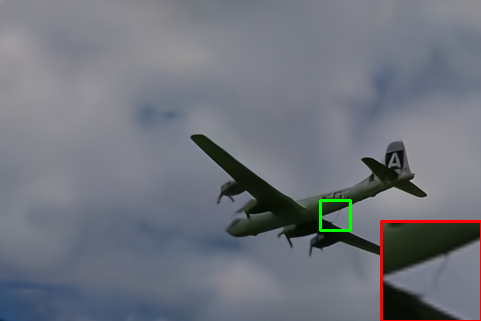}
        \caption*{0.9640}
    \end{subfigure}
    \begin{subfigure}{\subfigwidthreal}
        \includegraphics[width=\linewidth]{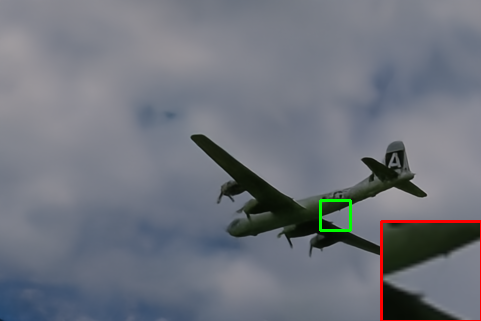}
        \caption*{0.9647}
    \end{subfigure}
    \begin{subfigure}{\subfigwidthreal}
        \includegraphics[width=\linewidth]{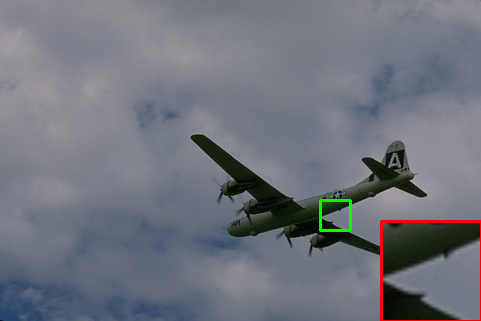}
        \caption*{SSIM}
    \end{subfigure}

    \begin{subfigure}{\subfigwidthreal}
        \includegraphics[width=\linewidth]{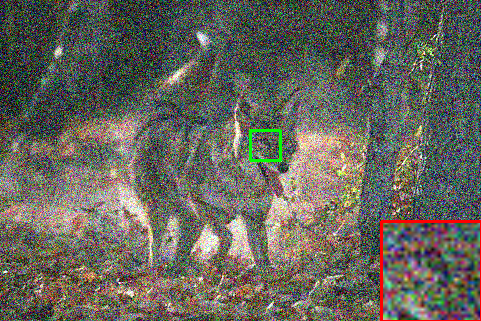}
        \caption*{Noisy}
    \end{subfigure}
    \begin{subfigure}{\subfigwidthreal}
        \includegraphics[width=\linewidth]{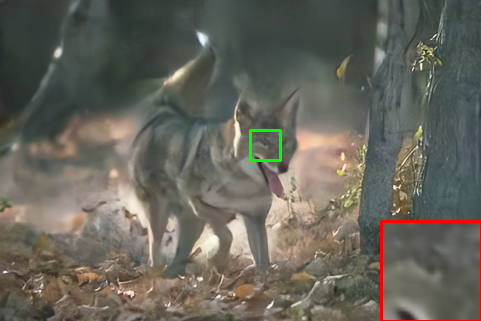}
        \caption*{0.8046}
    \end{subfigure}
    \begin{subfigure}{\subfigwidthreal}
        \includegraphics[width=\linewidth]{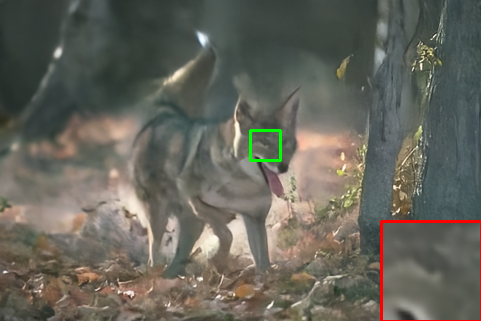}
        \caption*{0.8060}
    \end{subfigure}
    \begin{subfigure}{\subfigwidthreal}
        \includegraphics[width=\linewidth]{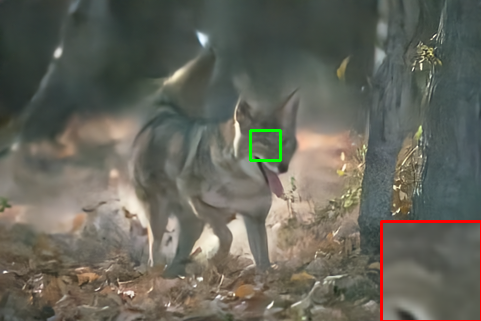}
        \caption*{0.7983}
    \end{subfigure}
    \begin{subfigure}{\subfigwidthreal}
        \includegraphics[width=\linewidth]{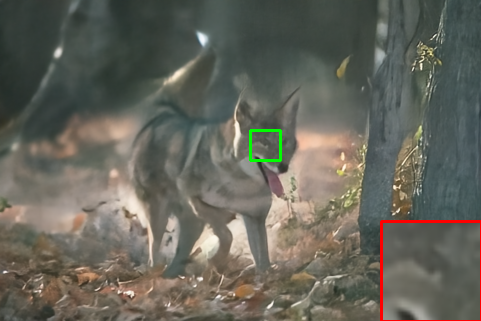}
        \caption*{0.8050}
    \end{subfigure}
    \begin{subfigure}{\subfigwidthreal}
        \includegraphics[width=\linewidth]{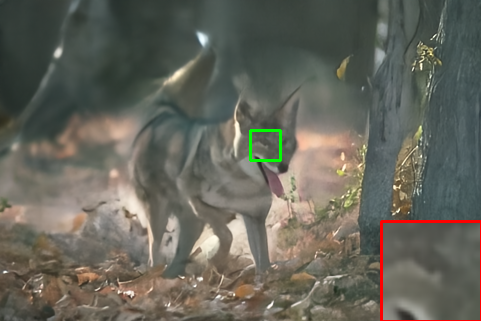}
        \caption*{0.8065}
    \end{subfigure}
    \begin{subfigure}{\subfigwidthreal}
        \includegraphics[width=\linewidth]{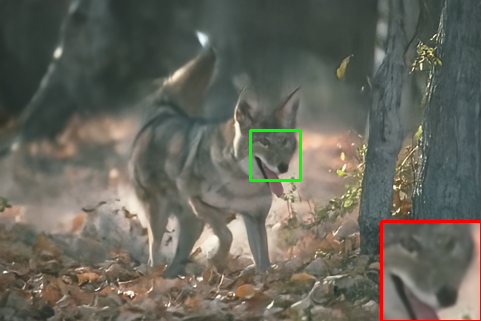} 
        \caption*{0.8397}
    \end{subfigure}
    \begin{subfigure}{\subfigwidthreal}
        \includegraphics[width=\linewidth]{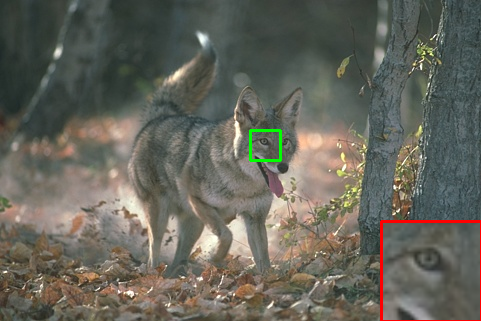}
        \caption*{SSIM}
    \end{subfigure}

    \vspace{-0.3cm}

    \begin{subfigure}{\subfigwidthreal}
        \caption*{Image}
    \end{subfigure}
    \begin{subfigure}{\subfigwidthreal}
        \caption*{MWDCNN}
    \end{subfigure}
    \begin{subfigure}{\subfigwidthreal}
        \caption*{DeamNet}
    \end{subfigure}
    \begin{subfigure}{\subfigwidthreal}
        \caption*{SwinIR}
    \end{subfigure}
    \begin{subfigure}{\subfigwidthreal}
        \caption*{MSANet}
    \end{subfigure}
    \begin{subfigure}{\subfigwidthreal}
        \caption*{DRANet}
    \end{subfigure}
    \begin{subfigure}{\subfigwidthreal}
        \caption*{MADNet}
    \end{subfigure}
    \begin{subfigure}{\subfigwidthreal}
        \caption*{Reference}
    \end{subfigure}
    \vspace{-3mm}
    \caption{Qualitative results on AWGN color images from the CBSD68 dataset. From left to right, we show the real noisy image, the results of MWDCNN, DeamNet, MSANet, DRANet, and MADNet.}
    \label{SIDD_color.png}
\end{figure}

For the visual comparisons shown in Figure~\ref{SIDD_color.png}, existing methods result in residual noises and pseudo artifacts. In contrast, our method recovers textures and structures more subtly and obtains clearer restoration. 

\subsection{Results on AWGN gray image denoising}
\label{Results on AWGN gray image Denoising}

\begin{table}[ht]
\centering
\scalebox{0.58}{
\begin{tabular}{c l c c c c c c c c c c c c}
\toprule
\multirow{2}{*}{\textbf{Datasets}} & \multirow{2}{*}{\textbf{Methods}} 
& \multicolumn{3}{c}{$\sigma = 15$} 
& \multicolumn{3}{c}{$\sigma = 25$} 
& \multicolumn{3}{c}{$\sigma = 30$}
& \multicolumn{3}{c}{$\sigma = 50$} \\
\cmidrule(lr){3-5}\cmidrule(lr){6-8}\cmidrule(lr){9-11}\cmidrule(lr){12-14}
& & \textbf{PSNR} & \textbf{SSIM} & \textbf{LPIPS} & \textbf{PSNR} & \textbf{SSIM} & \textbf{LPIPS} & \textbf{PSNR} & \textbf{SSIM} & \textbf{LPIPS} & \textbf{PSNR} & \textbf{SSIM} & \textbf{LPIPS} \\
\midrule
\multirow{8}{*}{BSD68}
 & DudeNet\cite{DudeNet}   & 29.27 & 0.779 & 0.1794 & 28.20 & 0.743 & 0.2295 & 27.28 & 0.703 & 0.2363 & 18.54 & 0.303 & 0.7034 \\
 & DeamNet\cite{DeamNet}   & 31.65 & 0.888 & 0.1200 & 29.25 & 0.827 & 0.1975 & 28.46 & 0.802 & 0.2259 & 26.41 & 0.728 & 0.3052 \\
 & SwinIR\cite{SwinIR}     & 31.69 & 0.890 & 0.1140 & 29.27 & 0.829 & 0.1804 & 28.47 & 0.803 & 0.2082 & 26.38 & 0.727 & 0.2900 \\
 & MSANet\cite{MSANet}    & 31.72 & 0.891 & 0.1075 & 29.30 & 0.831 & 0.1697 & 28.50 & 0.806 & 0.1984 & 26.42 & 0.729 & 0.2856 \\
 & AirNet\cite{AirNet}    & 31.65 & 0.890 & 0.1220 & 29.19 & 0.829 & 0.1850 & 28.40 & 0.800 & 0.2100 & 26.26 & 0.724 & 0.3020 \\
 & MWDCNN\cite{MWDCNN}    & 31.67 & 0.888 & 0.1240 & 29.24 & 0.827 & 0.1934 & 28.43 & 0.801 & 0.2229 & 26.33 & 0.724 & 0.2900 \\
 & DRANet\cite{DRANet}    & 31.78 & 0.891 & 0.1055 & 29.35 & 0.832 & 0.1665 & 28.55 & 0.807 & 0.1938 & 26.46 & 0.730 & 0.2772 \\
 & MADNet     & \textbf{31.81} & \textbf{0.893} & \textbf{0.0992} & \textbf{29.38} & \textbf{0.833} & \textbf{0.1631} & \textbf{28.58} & \textbf{0.808} & \textbf{0.1914} & \textbf{26.50} & \textbf{0.733} & \textbf{0.2615} \\
\midrule
\multirow{8}{*}{Set12}
 & DudeNet\cite{DudeNet}   & 29.98 & 0.779 & 0.1603 & 29.20 & 0.772 & 0.1680 & 28.11 & 0.726 & 0.1927 & 18.57 & 0.295 & 0.7263 \\
 & DeamNet\cite{DeamNet}   & 32.84 & 0.903 & 0.1219 & 30.56 & 0.865 & 0.1733 & 29.77 & 0.850 & 0.1896 & 27.54 & 0.798 & 0.2373 \\
 & SwinIR\cite{SwinIR}     & 32.86 & 0.903 & 0.1180 & 30.56 & 0.865 & 0.1570 & 29.77 & 0.850 & 0.1780 & 27.47 & 0.795 & 0.2191 \\
 & MSANet\cite{MSANet}    & 32.90 & 0.905 & 0.1115 & 30.61 & 0.867 & 0.1575 & 29.80 & 0.851 & 0.1799 & 27.55 & 0.797 & 0.2285 \\
 & AirNet\cite{AirNet}    & 32.75 & 0.902 & 0.1250 & 30.39 & 0.863 & 0.1700 & 29.70 & 0.847 & 0.1880 & 27.23 & 0.789 & 0.2300 \\
 & MWDCNN\cite{MWDCNN}    & 32.82 & 0.902 & 0.1233 & 30.53 & 0.864 & 0.1644 & 29.71 & 0.848 & 0.1801 & 27.40 & 0.791 & 0.2244 \\
 & DRANet\cite{DRANet}    & 32.99 & 0.904 & 0.1100 & 30.69 & 0.868 & 0.1555 & 29.88 & 0.852 & 0.1753 & 27.63 & 0.795 & 0.2203 \\
 & MADNet     & \textbf{33.00} & \textbf{0.905} & \textbf{0.1078} & \textbf{30.70} & \textbf{0.867} & \textbf{0.1549} & \textbf{29.90} & \textbf{0.852} & \textbf{0.1723} & \textbf{27.64} & \textbf{0.799} & \textbf{0.2189} \\
\bottomrule
\end{tabular}
}
\caption{Quantitative results of the AWGN gray noise removal evaluation for different noise levels $\sigma$.}
\label{tab:synthetic_comparison_gray}
\end{table}

Table~\ref{tab:synthetic_comparison_gray} presents the quantitative results on  on the gray images.
Our method consistently achieves the highest PSNR and SSIM across all noise levels tested on both datasets, demonstrating its strong denoising capability and robustness. In BSD68, our approach attains a PSNR of 31.81 dB at $\sigma$ = 15 and 26.50 dB at $\sigma$ = 50, effectively preserving structural details even at higher noise levels. Similarly, on Set12, our method maintains leading PSNR values of 33.05 dB at $\sigma$ = 15 and an SSIM of 0.801 at $\sigma$ = 50, underscoring its superior performance under varying noise conditions.

\begin{figure}[t!]
    \centering
    \begin{subfigure}{\subfigwidthreal}
        \includegraphics[width=\linewidth]{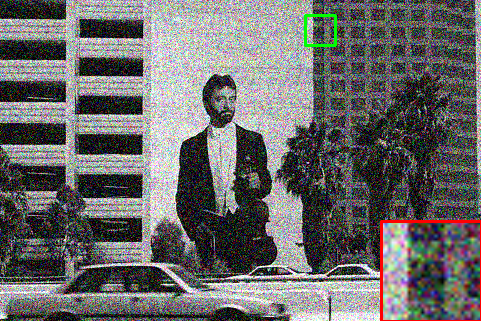}
        \caption*{Noisy}
    \end{subfigure}
    \begin{subfigure}{\subfigwidthreal}
        \includegraphics[width=\linewidth]{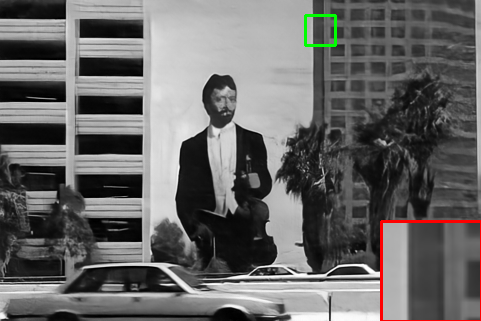}
        \caption*{0.7838}
    \end{subfigure}
    \begin{subfigure}{\subfigwidthreal}
        \includegraphics[width=\linewidth]{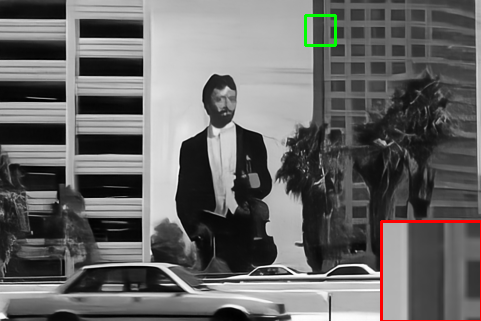}
        \caption*{0.7955}
    \end{subfigure}
    \begin{subfigure}{\subfigwidthreal}
        \includegraphics[width=\linewidth]{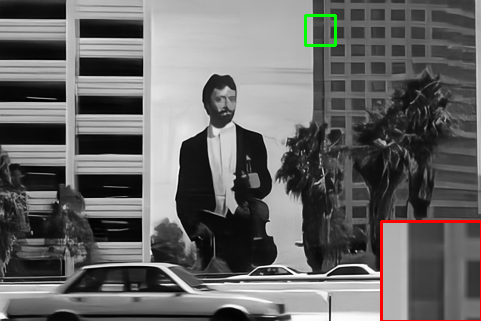}
        \caption*{0.8301}
    \end{subfigure}
    \begin{subfigure}{\subfigwidthreal}
        \includegraphics[width=\linewidth]{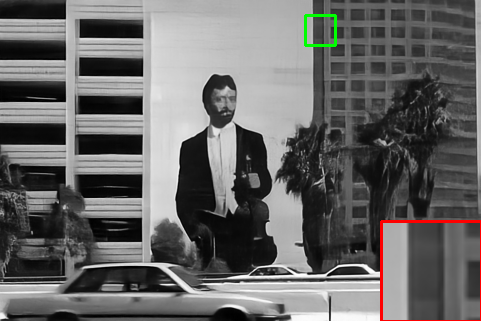}
        \caption*{0.7957}
    \end{subfigure}
    \begin{subfigure}{\subfigwidthreal}
        \includegraphics[width=\linewidth]{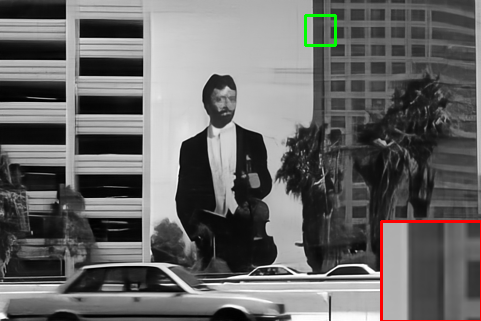}
        \caption*{0.7990}
    \end{subfigure}
    \begin{subfigure}{\subfigwidthreal}
        \includegraphics[width=\linewidth]{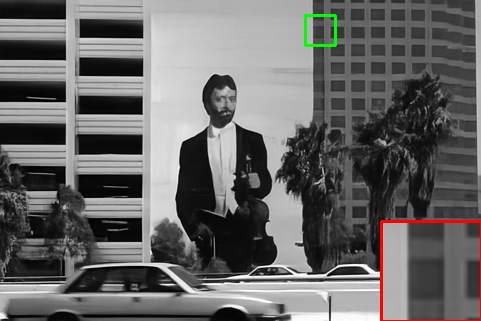}
        \caption*{0.8045}
    \end{subfigure}
    \begin{subfigure}{\subfigwidthreal}
        \includegraphics[width=\linewidth]{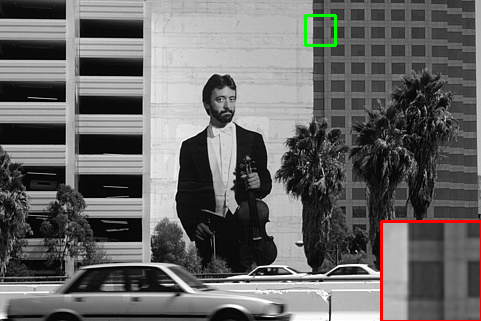}
        \caption*{SSIM}
    \end{subfigure}

    \begin{subfigure}{\subfigwidthreal}
        \includegraphics[width=\linewidth]{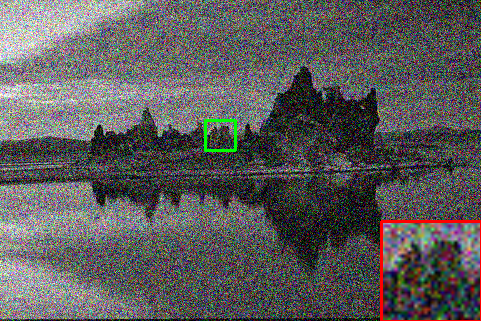}
        \caption*{Noisy}
    \end{subfigure}
    \begin{subfigure}{\subfigwidthreal}
        \includegraphics[width=\linewidth]{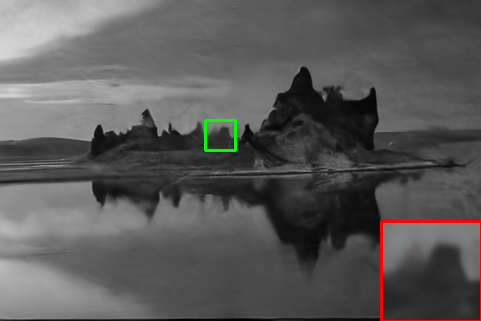}
        \caption*{0.8318}
    \end{subfigure}
    \begin{subfigure}{\subfigwidthreal}
        \includegraphics[width=\linewidth]{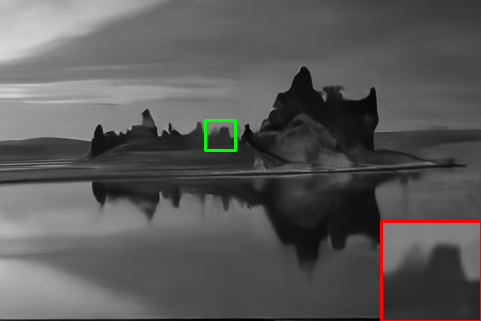}
        \caption*{0.8390}
    \end{subfigure}
    \begin{subfigure}{\subfigwidthreal}
        \includegraphics[width=\linewidth]{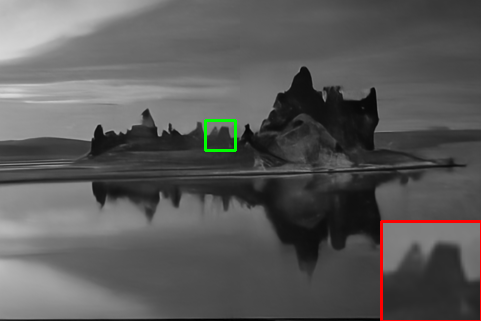}
        \caption*{0.8564}
    \end{subfigure}
    \begin{subfigure}{\subfigwidthreal}
        \includegraphics[width=\linewidth]{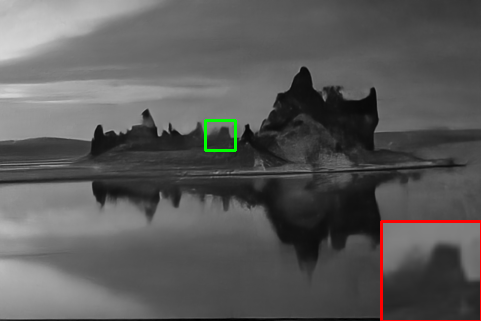}
        \caption*{0.8417}
    \end{subfigure}
    \begin{subfigure}{\subfigwidthreal}
        \includegraphics[width=\linewidth]{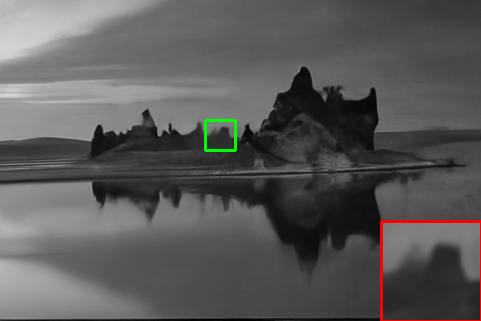}
        \caption*{0.8419}
    \end{subfigure}
    \begin{subfigure}{\subfigwidthreal}
        \includegraphics[width=\linewidth]{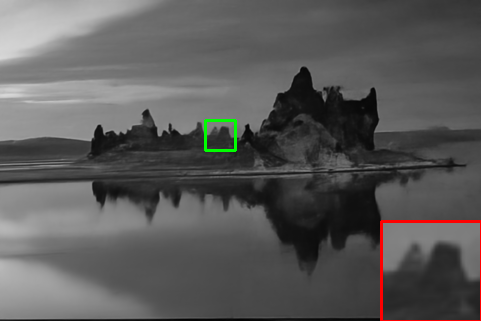}
        \caption*{0.8628}
    \end{subfigure}
    \begin{subfigure}{\subfigwidthreal}
        \includegraphics[width=\linewidth]{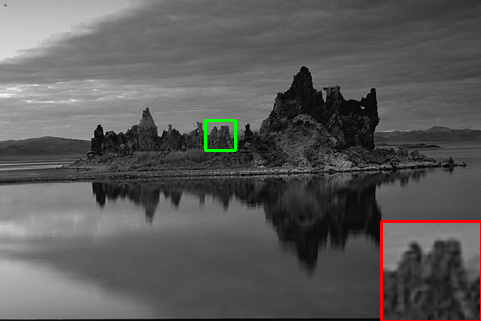}
        \caption*{SSIM}
    \end{subfigure}

    \vspace{-0.3cm}

    \begin{subfigure}{\subfigwidthreal}
        \caption*{Image}
    \end{subfigure}
    \begin{subfigure}{\subfigwidthreal}
        \caption*{MWDCNN}
    \end{subfigure}
    \begin{subfigure}{\subfigwidthreal}
        \caption*{DeamNet}
    \end{subfigure}
    \begin{subfigure}{\subfigwidthreal}
        \caption*{SwinIR}
    \end{subfigure}
    \begin{subfigure}{\subfigwidthreal}
        \caption*{MSANet}
    \end{subfigure}
    \begin{subfigure}{\subfigwidthreal}
        \caption*{DRANet}
    \end{subfigure}
    \begin{subfigure}{\subfigwidthreal}
        \caption*{MADNet}
    \end{subfigure}
    \begin{subfigure}{\subfigwidthreal}
        \caption*{Reference}
    \end{subfigure}
    \vspace{-3mm}
    \caption{Qualitative results on AWGN gray images from the CBSD68 dataset. From left to right, we show the real noisy image, the results of MWDCNN, DeamNet, MSANet, DRANet, and MADNet.}
    \label{SIDD_gray.png}
\end{figure}

For the visual comparisons shown in Figure~\ref{SIDD_gray.png}, our approach preserves textures and structures more delicately, achieving a sharper and clearer restoration. 

Overall, these results validate the effectiveness and generalizability of our method for image denoising across diverse datasets and noise levels, as it consistently achieves the best quantitative results without sacrificing image quality in higher noise regimes.

\subsection{Results on real-image denoising}
\label{Results on real-image denoising}

We also compared MADNet with several state-of-the-art image denoising methods (DeamNet \cite{DeamNet}, VDIR \cite{VDIR}, MSANet \cite{MSANet}, CFMNet \cite{CFMNet}, and DRANet \cite{DRANet}) on two benchmark datasets (SIDD and DND), and evaluated their performance using PSNR and SSIM metrics.
We use the corresponding source codes provided by their authors,
and adhere strictly to the identical experimental settings across all evaluations.

\begin{table}[t!]
\centering
\tabcolsep=0.6cm
\begin{tabular}{c c c c}
\toprule
\textbf{Datasets} & \textbf{Methods} & \textbf{PSNR} & \textbf{SSIM} \\
\midrule
\multirow{6}{*}{SIDD} 
 & DeamNet\cite{DeamNet} & 39.35 & 0.955 \\
 & VDIR\cite{VDIR}    & 39.26 & 0.955 \\
 & MSANet\cite{MSANet}  & 39.26 & 0.955 \\
 & CFMNet\cite{CFMNet}  & 39.24 & 0.954 \\
 & DRANet\cite{DRANet}  & 39.54 & 0.957 \\
 & \textbf{MADNet} & \textbf{39.78} & \textbf{0.959} \\
\midrule
\multirow{6}{*}{DND}
 & DeamNet\cite{DeamNet} & 39.63 & 0.953 \\
 & VDIR\cite{VDIR}    & 39.63 & 0.953 \\
 & MSANet\cite{MSANet}  & 39.25 & 0.952 \\
 & CFMNet\cite{CFMNet}  & 38.86 & 0.947 \\
 & DRANet\cite{DRANet}  & 39.63 & 0.953 \\
 & \textbf{MADNet} & \textbf{39.98} & \textbf{0.956} \\
\bottomrule

\end{tabular}
\caption{Quantitative results (PSNR and SSIM) of the real noise removal evaluation.}
\label{tab:real noise}
\end{table}

Table~\ref{tab:real noise} presents the performance of our real-image denoising method. On SIDD dataset \cite{SIDD}, our approach achieves an impressive PSNR of 39.78 dB and SSIM of 0.959. On DND dataset \cite{DND}, our method attains a PSNR of 39.98 dB and SSIM of 0.956. These results demonstrate the superior denoising capability and strong generalization ability of our approach across two standard real-world datasets.

\begin{figure}[t!]
    \centering
    \begin{subfigure}{\subfigwidthreal}
        \includegraphics[width=\linewidth]{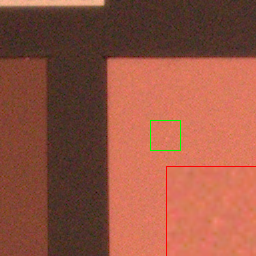}
        \caption*{Noisy}
    \end{subfigure}
    \begin{subfigure}{\subfigwidthreal}
        \includegraphics[width=\linewidth]{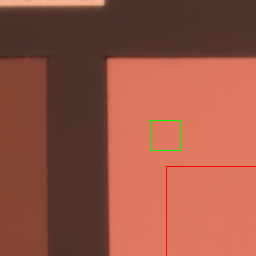}
        \caption*{0.9843}
    \end{subfigure}
    \begin{subfigure}{\subfigwidthreal}
        \includegraphics[width=\linewidth]{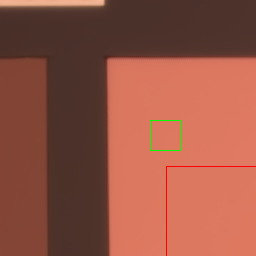}
        \caption*{0.9851}
    \end{subfigure}
    \begin{subfigure}{\subfigwidthreal}
        \includegraphics[width=\linewidth]{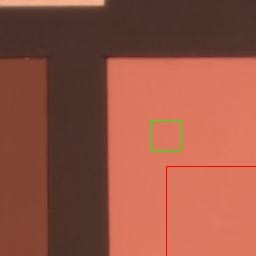}
        \caption*{0.9850}
    \end{subfigure}
    \begin{subfigure}{\subfigwidthreal}
        \includegraphics[width=\linewidth]{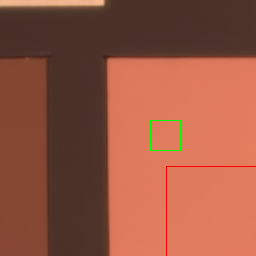}
        \caption*{0.9851}
    \end{subfigure}
    \begin{subfigure}{\subfigwidthreal}
        \includegraphics[width=\linewidth]{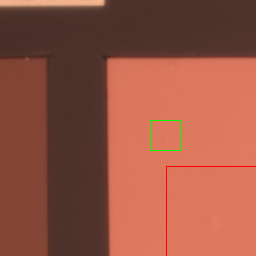}
        \caption*{0.9858}
    \end{subfigure}
    \begin{subfigure}{\subfigwidthreal}
        \includegraphics[width=\linewidth]{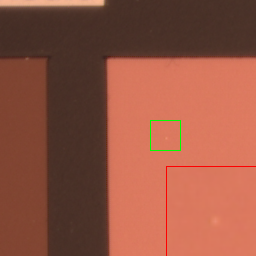}
        \caption*{SSIM}
    \end{subfigure}

    \begin{subfigure}{\subfigwidthreal}
        \includegraphics[width=\linewidth]{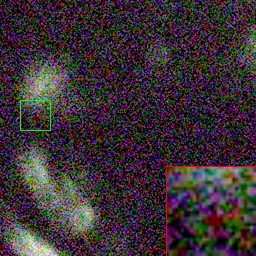}
        \caption*{Noisy}
    \end{subfigure}
    \begin{subfigure}{\subfigwidthreal}
        \includegraphics[width=\linewidth]{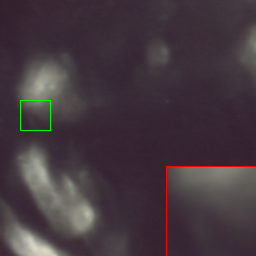}
        \caption*{0.8644}
    \end{subfigure}
    \begin{subfigure}{\subfigwidthreal}
        \includegraphics[width=\linewidth]{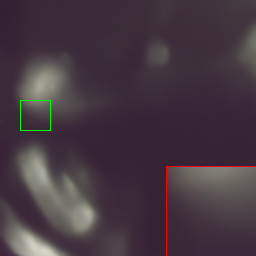}
        \caption*{0.8662}
    \end{subfigure}
    \begin{subfigure}{\subfigwidthreal}
        \includegraphics[width=\linewidth]{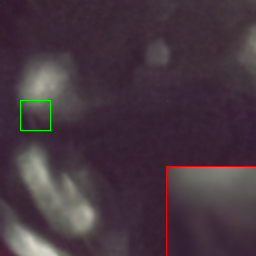}
        \caption*{0.8596}
    \end{subfigure}
    \begin{subfigure}{\subfigwidthreal}
        \includegraphics[width=\linewidth]{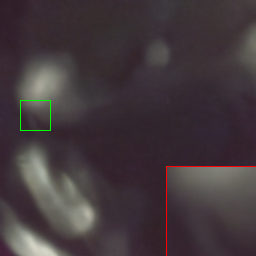}
        \caption*{0.8630}
    \end{subfigure}
    \begin{subfigure}{\subfigwidthreal}
        \includegraphics[width=\linewidth]{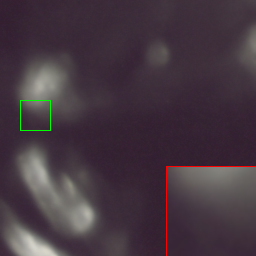}
        \caption*{0.8672}
    \end{subfigure}
    \begin{subfigure}{\subfigwidthreal}
        \includegraphics[width=\linewidth]{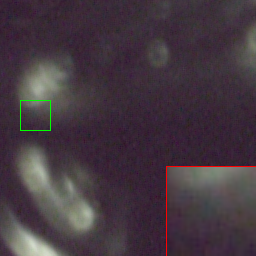}
        \caption*{SSIM}
    \end{subfigure}

    \vspace{-0.3cm}

    \begin{subfigure}{\subfigwidthreal}
        \caption*{Image}
    \end{subfigure}
    \begin{subfigure}{\subfigwidthreal}
        \caption*{DeamNet}
    \end{subfigure}
    \begin{subfigure}{\subfigwidthreal}
        \caption*{MSANet}
    \end{subfigure}
    \begin{subfigure}{\subfigwidthreal}
        \caption*{CFMNet}
    \end{subfigure}
    \begin{subfigure}{\subfigwidthreal}
        \caption*{DRANet}
    \end{subfigure}
    \begin{subfigure}{\subfigwidthreal}
        \caption*{MADNet}
    \end{subfigure}
    \begin{subfigure}{\subfigwidthreal}
        \caption*{Reference}
    \end{subfigure}
    
    \caption{Qualitative results on real noisy images from the SIDD validation dataset. From left to right, we show the real noisy image, the results of DeamNet, MSANet, CFMNet, DRANet, and MADNet.}
    \label{SIDD_real.png}
\end{figure}

For the visual comparisons shown in Figure \ref{SIDD_real.png}, existing works destroy the textures and obtain over-smoothed results. 
In contrast, our approach better preserves structural details and textural patterns, yielding sharper and more faithful reconstructions.
Key regions are highlighted with colored rectangles, and zooming in is advised for improved clarity.

\subsection{Ablation Study}
\label{Ablation Study}

To present the effectiveness of our MADNet, we conduct various ablation
studies about multi-scale framework, loss function and the dual-domain modulation layer. The evaluations are conducted on the SIDD Medium
dataset \cite{SIDD}.
Table~\ref{tab:ablation} presents the results of the ablation study of our denoising framework, highlighting the performance of PSNR (dB) when various components of the multi-scale framework and the dual-domain modulation layers are removed or retained.

\begin{table}[t!]
    \centering
    \resizebox{1.0\textwidth}{!}{%
    \begin{tabular}{l|cccc|cccc|c}
    \toprule
    \multirow{2}{*}{\textbf{Method}} 
    & \multicolumn{4}{c|}{\textbf{Multi-scale Framework}} 
    & \multicolumn{4}{c|}{\textbf{Dual-Domain Modulation Layer}} 
    & \multirow{2}{*}{\textbf{PSNR (dB)}} \\
    \cline{2-9}
     & MSI & SF & MSL & MFL & ASEB & AFEB & Sep & Enh & \\
    \midrule
    \textbf{Full Model}        & \cmark & \cmark & \cmark & \cmark & \cmark & \cmark & \cmark & \cmark & \textbf{39.777}\\
    \midrule
    Replace TSA with W-MSA & \cmark & \cmark & \cmark & \cmark & \cmark & \cmark & \cmark & \cmark & 39.720 \\
    \midrule
    w/o MSI              & \xmark & \cmark & \cmark & \cmark & \cmark & \cmark & \cmark & \cmark & 39.705 \\
    w/o GFFB               & \cmark & \xmark & \cmark & \cmark & \cmark & \cmark & \cmark & \cmark & 39.714 \\
    w/o MSI \& GFFB        & \xmark & \xmark & \cmark & \cmark & \cmark & \cmark & \cmark & \cmark & 39.695 \\
    w/o MSL              & \cmark & \cmark & \xmark & \cmark & \cmark & \cmark & \cmark & \cmark & 39.734 \\
    w/o MFL              & \cmark & \cmark & \cmark & \xmark & \cmark & \cmark & \cmark & \cmark & 39.733 \\
    \midrule
    w/o ASEB              & \cmark & \cmark & \cmark & \cmark & \xmark & \cmark & \cmark & \cmark & 39.663 \\
    w/o AFEB             & \cmark & \cmark & \cmark & \cmark & \cmark & \xmark & \cmark & \cmark & 39.696 \\
    w/o ASEB and AFEB    & \cmark & \cmark & \cmark & \cmark & \xmark & \xmark & \cmark & \cmark & 39.603 \\
    AFEB w/o Sep         & \cmark & \cmark & \cmark & \cmark & \xmark & \xmark & \xmark & \cmark & 39.625 \\
    AFEB w/o Enh         & \cmark & \cmark & \cmark & \cmark & \xmark & \xmark & \cmark & \xmark & 39.621 \\
    ASEB + AFEB w/o Sep   & \cmark & \cmark & \cmark & \cmark & \cmark & \cmark & \xmark & \cmark & 39.670 \\
    ASEB + AFEB w/o Enh   & \cmark & \cmark & \cmark & \cmark & \cmark & \cmark & \cmark & \xmark & 39.717 \\
    \bottomrule
    \end{tabular}
    }
    \vspace{4pt}
    {\footnotesize
    \textbf{MSI}: Multi-scale Input,\,
    \textbf{SF}: Spatial-Frequency Fusion,\,
    \textbf{MSL}: Multi-scale Spatial Loss,\,
    \textbf{MFL}: Multi-scale Frequency Loss,\,
    \textbf{ASEB}: Adaptive Spatial Enhancement Block,\,
    \textbf{AFEB}: Adaptive Frequency Enhancement Block,\,
    \textbf{Sep}: Separation,\,
    \textbf{Enh}: Enhancement,\,
    \textbf{TSA}: Transposed Self-Attention,\,
    \textbf{W-MSA}: Window-based Multi-head Self-Attention
    }
    \caption{Ablation study results on the SIDD dataset. PSNR is reported in decibels (dB).}
    \label{tab:ablation}
\end{table}

\subsubsection{Ablation study with multi-scale framework}

Table~\ref{tab:ablation} demonstrates both multi-scale input and global feature fusion block (GFFB) are essential for high performance. 
Removing either component causes a minor PSNR drop, while removing both results in a more noticeable 0.04 dB reduction (39.695 dB vs 39.735 dB). 
Removing the multiscale spatial or frequency-domain loss affects performance (39.734 dB and 39.733 dB), as they enhance training stability and convergence.

\subsubsection{Ablation study with dual-domain modulation layers}

Focusing on the dual-domain modulation layer, using only the spatial module (39.663 dB) or frequency module (39.696 dB) results in lower performance, highlighting that single-domain reconstruction is insufficient. 
Removal of frequency separation (39.670 dB) or frequency enhancement (39.621–39.625 dB) significantly degrades performance, underscoring their importance in the frequency domain.
Including all components of the multiscale framework and the dual-domain modulation layer achieves the highest PSNR (39.777 dB), confirming the complementary benefits of these modules in our denoising network.

\subsection{Limitation}
\label{Model complexity analysis}

\begin{table}[t!]
    \centering
    \tabcolsep=0.6cm
    \begin{tabular}{lccc}
        \toprule
        \textbf{Model Name} & \textbf{FLOPs (G)} & \textbf{Parameters (M)}  & \textbf{Time (s)} \\
        \midrule
        DeamNet\cite{DeamNet} & 145.8 & 1.876 & 0.0125 \\
        DudeNet\cite{DudeNet} & 70.87 & 1.079 & 0.0031 \\
        MSANet\cite{MSANet}  & 35.35 & 7.997 & 0.0096 \\
        CFMNet\cite{CFMNet}  & 238.3 & 22.05 & 0.0175 \\
        DRANet\cite{DRANet}  & 592.3 & 1.617 & 0.0225 \\
        MADNet  & 91.35 & 28.38 & 0.0482 \\
        \bottomrule
    \end{tabular}
    \caption{Comparison of FLOPs (G) and Parameters (M) of Various Models at $256 \times 256$ Resolution}
    \label{tab:flops_parameters_comparison}
\end{table}

In this section, we compare the Floating-Point Operations (FLOPs) and parameters of six denoising models: DeamNet \cite{DeamNet}, DudeNet \cite{DudeNet}, MSANet \cite{MSANet}, CFMNet \cite{CFMNet}, DRANet \cite{DRANet}, and our proposed MADNet on image resolution of $256 \times 256$.
Table \ref{tab:flops_parameters_comparison} demonstrates that our model achieves competitive performance in terms of Floating-Point Operations (FLOPs), but with a relatively large number of parameters. 

\section{Conclusion}
\label{Conclusion}

In this paper, we propose an adaptive spatial-frequency learning unit (ASFU) and build a novel multi-scale adaptive dual-domain network (MADNet) for image denoising. 
We observed that different types of degradation in frequency domain can affect the image content in different degrees. Motivated by this, ASFU uses a learnable mask and transposed self-attention to separate and realize the interaction of high-frequency and low-frequency information.
We also use low-resolution images to provide enriched information and guide the image denoising process. 
In addition, we design GFFB to enhance the features at different scales.
Extensive experiments demonstrate that MADNet outperforms current state-of-the-art denoising approaches.

However, we mainly concentrate on the effectiveness and the performance in terms of FLOPs in this paper, further research is needed to reduce the number of parameters of MADNet while maintaining its performance in denoising tasks. 
Furthermore, we expect to explore the possibility of implementing MADNet for more general image restoration tasks.

\bibliographystyle{elsarticle-num} 
\bibliography{elsarticle-template-num.bib}

\end{document}